\pdfoutput=1

\documentclass[11pt]{article}

\usepackage{emnlp2021}

\usepackage{times}
\usepackage{latexsym}
\usepackage{url}
\usepackage{booktabs}
\usepackage{graphicx}
\usepackage{cjhebrew}
\usepackage{ucs}
\usepackage{tcolorbox}
\usepackage{wrapfig}
\usepackage{subcaption}
\usepackage{multirow}
\usepackage{leipzig}
\usepackage{expex}

\usepackage[T1]{fontenc}

\usepackage[utf8x]{inputenc}

\usepackage{microtype}

\title{\textsc{ParaShoot}: A Hebrew Question Answering Dataset}

\author{Omri Keren \qquad \qquad Omer Levy\\
  The Blavatnik School of Computer Science\\
  Tel Aviv University \\
  \texttt{omrikeren@mail.tau.ac.il}}

\begin{document}

\maketitle

\begin{abstract}
NLP research in Hebrew has largely focused on morphology and syntax, where rich annotated datasets in the spirit of Universal Dependencies are available.
Semantic datasets, however, are in short supply, hindering crucial advances in the development of NLP technology in Hebrew.
In this work, we present \textsc{ParaShoot}, the first question answering dataset in modern Hebrew.
The dataset follows the format and crowdsourcing methodology of SQuAD, and contains approximately 3000 annotated examples, similar to other question-answering datasets in low-resource languages.
We provide the first baseline results using recently-released BERT-style models for Hebrew, showing that there is significant room for improvement on this task.
\end{abstract}

\section{Introduction}

Natural language processing has seen a surge in the pretraining paradigm in recent years with the appearance of pretrained models in a plethora of languages, including Hebrew \citep{chriqui2021hebert, seker2021alephbert}. 
While such models have shown to perform remarkably well on a variety of tasks, most of the evaluation of the Hebrew models, however, has been focused on morphology and syntax tasks in the spirit of universal dependencies \citep{nivre-etal-2017-universal}, while end-user-focused evaluation has been limited to sentiment analysis \citep{chriqui2021hebert} and named entity recognition \citep{bareket2020neural}.

In this paper, we try to remedy the scarcity of semantic datasets by presenting \textsc{ParaShoot},\footnote{A portmanteau of \textit{paragraph} and \cjRL{t}"\cjRL{/sw} (\textit{shoot}), the Hebrew abbreviation of Q\&A.} the first question answering dataset in Hebrew, in the style of SQuAD \citep{rajpurkar-etal-2016-squad}.
We follow similar work in constructing non-English question answering datasets \cite[\textit{inter alia}]{dhoffschmidt-etal-2020-fquad, mozannar-etal-2019-neural, lim2019korquad1}, 
and turn to Hebrew-speaking crowdsource workers, asking them to write questions given paragraphs sampled at random from Hebrew Wikipedia. Through this process, we collect approximately 3000 annotated (\textit{paragraph}, \textit{question}, \textit{answer}) triplets, in a setting that may be suitable for few-shot learning, simulating the amount of data a startup or academic group can quickly collect with a limited annotation budget or a short deadline.

Statistical analysis of \textsc{ParaShoot} shows that the dataset is diverse in question types and complexity, and that the annotations are of decent quality.
We provide baseline results based on two recently-released BERT-style models in Hebrew, showing that there is much potential in devising better pretraining and fine-tuning schemes to improve the performance of Hebrew language models on this dataset.
We hope that this new dataset will pave the way for practitioners and researchers to advance natural language understanding in Hebrew.\footnote{The dataset is publicly available at \url{https://github.com/omrikeren/ParaShoot}}

\section{Dataset}

We present \textsc{ParaShoot}, a question answering dataset in Hebrew, in a format that closely follows that of SQuAD \citep{rajpurkar-etal-2016-squad}. Each example in the dataset is a triplet consisting of a paragraph, a question, and a span from the paragraph text constituting the answer to the question.
We scrape paragraphs from random Hebrew Wikipedia articles, and crowdsource questions and answers for each one, resulting in 3038 annotated examples.
While larger datasets may facilitate better-performing models,
recent work has advocated for research on smaller labeled datasets \cite{ram-etal-2021-shot}, which more accurately reflect the amount of data a startup or academic lab can collect in a short amount of time and resources.


\subsection{Corpus}

We collect random articles from Hebrew Wikipedia, covering a wide range of domains and topics. 
We only sample articles containing at least two paragraphs and 500 characters.\footnote{We filter out images, tables, etc.}
Finally, for each such article, two candidate paragraphs are randomly sampled and added to the annotation corpus.
These paragraphs will eventually become the passages in the question answering dataset.

\begin{figure}[t!]
    \centering
    \includegraphics[width=\columnwidth]{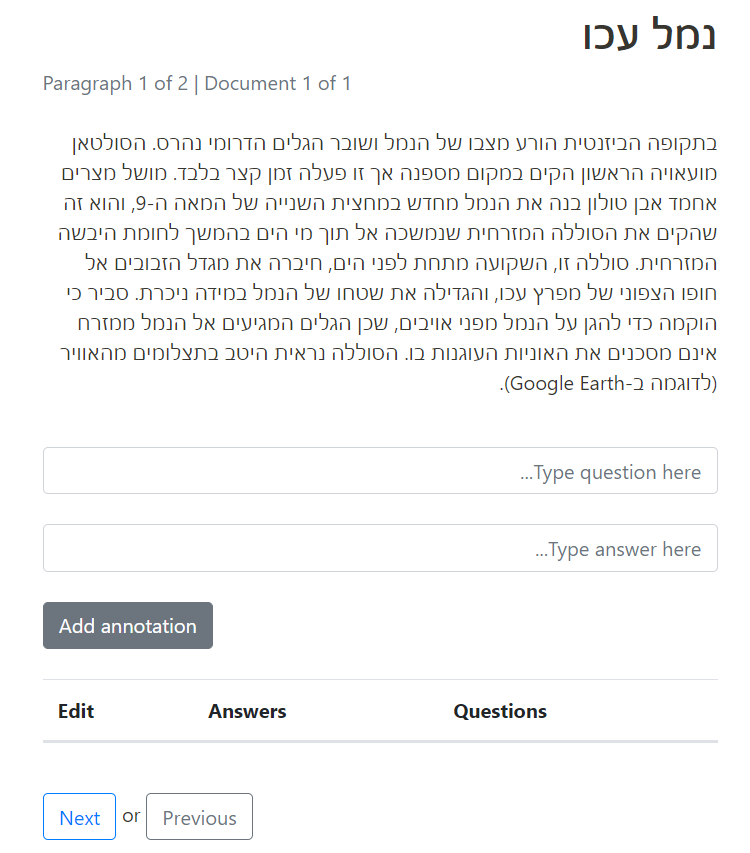}
    \caption{The annotation user interface, containing the article's title, the paragraph, a slot for entering a question, and an additional slot for entering the answer. Dragging the mouse over a span in the paragraph automatically fills the question slot, allowing for quick and accurate annotation of answer spans.}
    \label{fig:web-example}
\end{figure}

\subsection{Annotation}

We recruit annotators by using the Prolific crowdsourcing platform.\footnote{\url{www.prolific.co}}
Being a native Hebrew speaker is the only required qualification, allowing the participation of a few dozen annotators in the campaign.
Annotators are presented with random paragraphs from the annotation set, and tasked to write 3-5 questions that are explicitly answered by the given text, for each paragraph.
As in the original SQuAD annotation campaign, annotators are instructed to phrase the questions in their own words, and highlight the minimal span of characters from the paragraph that contains the answer to each question.
Our implementation also provides automatic data validation heuristics that alert the annotators if, for instance, the answer span is too long or not a substring of the paragraph. 
Figure~\ref{fig:web-example} shows a screenshot from the annotation web page.\footnote{The platform's code is based upon \url{https://github.com/cdqa-suite/cdQA-annotator}.}

We acknowledge the fact that this data collection technique is known to encourage annotation artifacts \cite{gururangan-etal-2018-annotation, kaushik-lipton-2018-much}, and several newer annotation methods, such as TyDi QA \cite{clark-etal-2020-tydi}, have been introduced to alleviate them. 
Nevertheless, we follow SQuAD's annotation methodology, as it necessitates considerably fewer resources.
Maintaining an hourly wage of over \$10,\footnote{7.50 GBP $\approx$ 10.50 USD, at the time of writing.} we were able to collect our entire dataset, including discarded data from development runs, for under \$800.

\subsection{Post-Processing}

In total, we amass 3106 question-answer examples. Of those, we discard 68 examples (2.2\%) that contained yes/no questions or extremely short/long answers.
The resulting dataset contains 3038 examples, which we divide to training, validation, and test by article, preventing content overlap.
Table~\ref{table:stats} details the amount of unique articles, paragraphs, and questions of each split.

\begin{table}[t!]
\small
\centering
\begin{tabular}{@{}lrrr@{}} 
\toprule   
 & \textbf{\#Articles} & \textbf{\#Paragraphs} & \textbf{\#Questions}  \\
\midrule
Train & 295 & 565  & 1792 \\ 
Validation & 33 & 63 & 221 \\
Test & 165 & 319  & 1025 \\
\midrule
Total & 493 & 947 & 3038 \\ 
\bottomrule
\end{tabular}
\caption{The number of unique articles, paragraphs, and questions in each split of \textsc{ParaShoot}. The dataset is partitioned by articles.}
\label{table:stats}
\end{table}

\section{Analysis}

We analyze the dataset in various ways to assess its quality and limitations as a benchmark.

\begin{figure*}[t]
     \centering
    \includegraphics[width=\columnwidth]{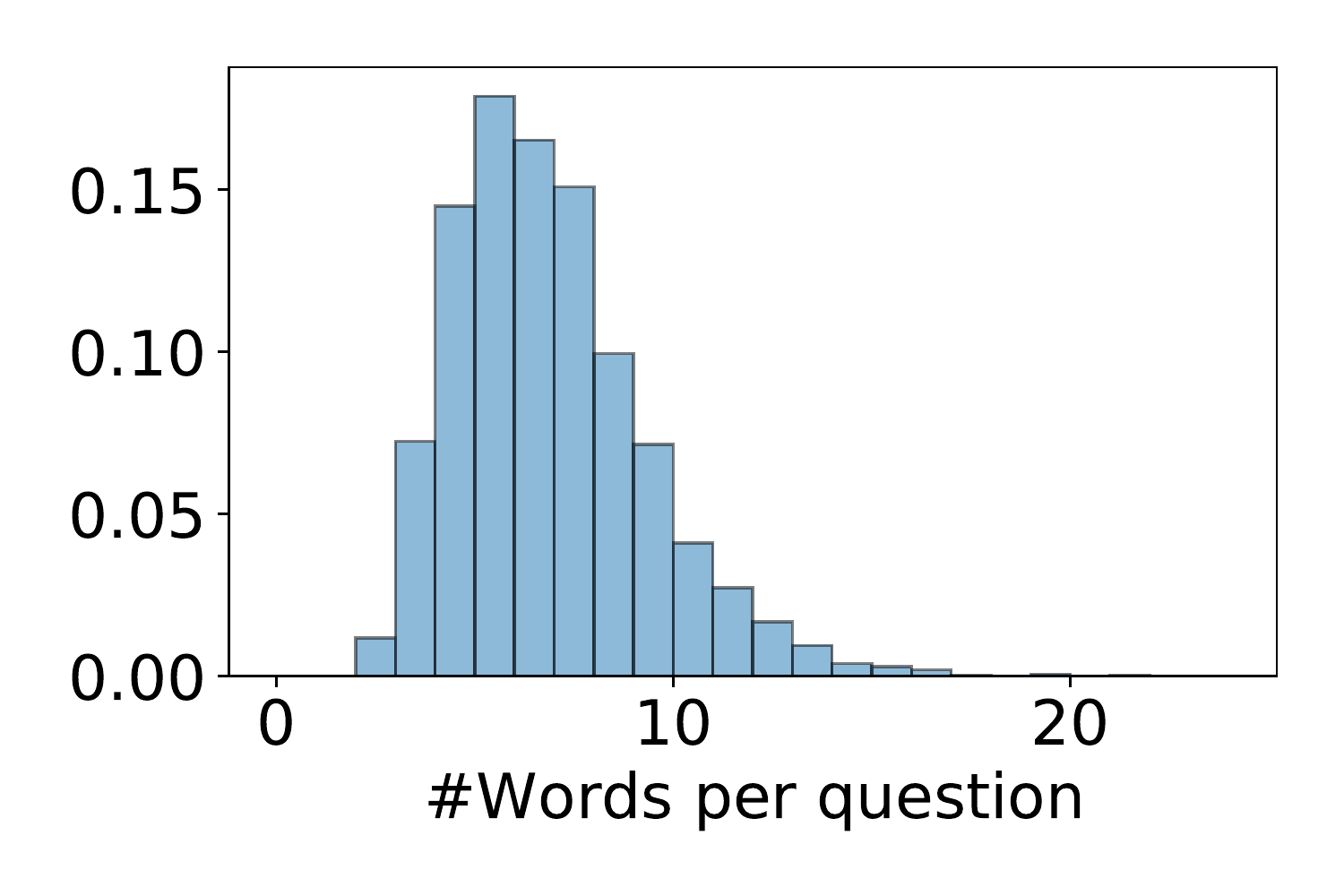}
     \hfill
    \includegraphics[width=\columnwidth]{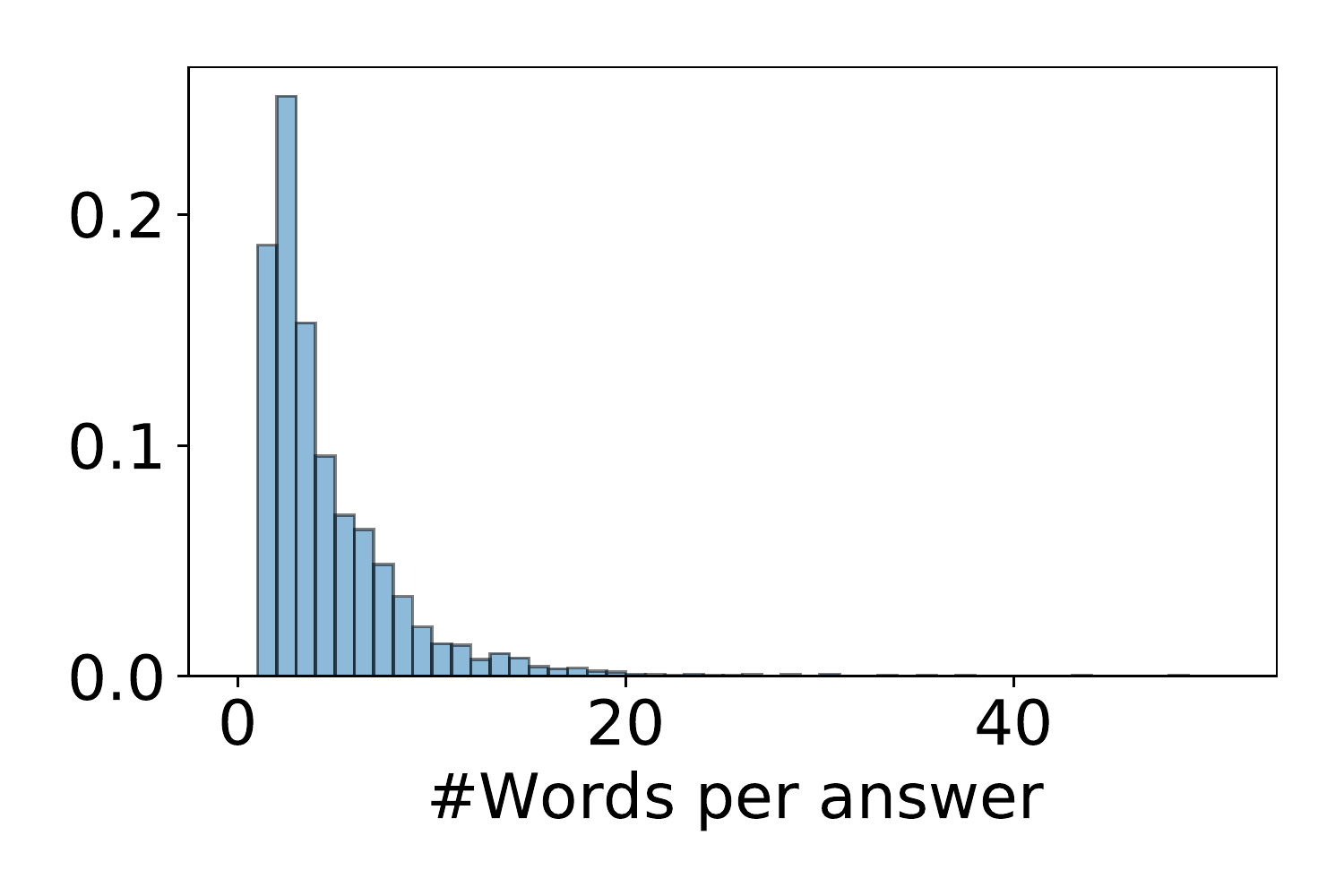}
\caption{The length distribution of questions (left) and answers (right) in the entire dataset.}
\label{fig:token-hist}
\end{figure*}

\subsection{Annotation Quality}
\label{sec:quality}

\begin{figure*}[h]
\centering
\includegraphics[width=\textwidth,trim=2cm 11.5cm 1.5cm 2.9cm,clip]{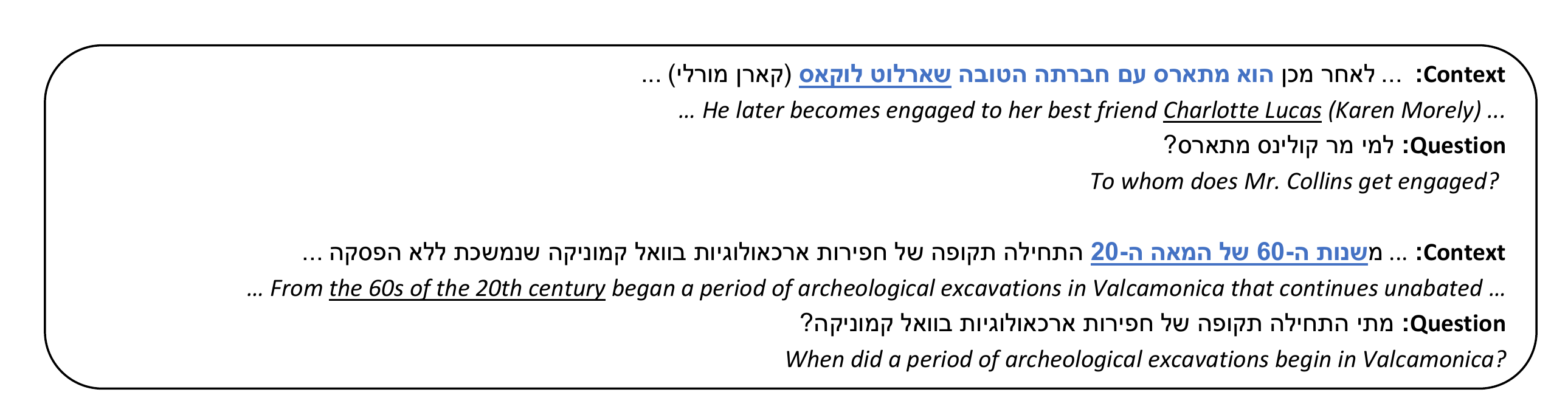}
\caption{Examples from the validation set. The text in bold shows crowd-annotated answers. The underlined text represents the (expert-annotated) minimal answer span. The first example demonstrates a non-minimal span that has some overlap with the question's text. The second example demonstrates a valid minimal span selection.}
\label{fig:annotation-example}
\end{figure*}

To measure the quality of the annotated data, we randomly select 50 examples from the validation set, and manually analyze them ourselves.\footnote{The authors are native speakers of modern Hebrew.}
Specifically, we check whether the annotated answer span is \textit{correct} (answers the question) and \textit{minimal} (contains only the answer).
Table~\ref{tab:data-quality} shows that the majority of the annotations are indeed valid, answering the questions with a minimal span.
Yet, a significant minority contains additional supporting information, which makes the answer span longer than the desired minimal span by 2.5 times on average.
We can thus expect an upper bound of 57\% token F1 on those examples, setting the performance ceiling at around 84\% F1 for the entire dataset.
Finally, we present examples from the validation set that illustrate the annotation quality (Figure~\ref{fig:annotation-example}). 

\begin{table}[t]
\small
\centering
\begin{tabular}{@{}lr@{}}
\toprule
\textbf{Answer Span} & \textbf{Frequency} \\
\midrule
Minimal & 70\% \\
Too Long & 28\%  \\
Too Short & 2\% \\
\bottomrule
\end{tabular}
\caption{Distribution of annotated answer span quality, based on manual analysis of 50 examples from the validation set.}
\label{tab:data-quality}
\end{table}


\begin{table}[t!]
\small
\centering
\begin{tabular}{@{}lrr@{}} 
\toprule   
\multicolumn{2}{@{}l}{\textbf{Question Word}} & \textbf{Frequency} \\
\midrule
What & \dots/\cjRL{mhw}/\cjRL{mh} & 16.29\% \\ 
Which & \dots/\cjRL{'yzw}/\cjRL{'yzh}  & 15.84\% \\
Who & \dots/\cjRL{myhw}/\cjRL{my} & 14.03\% \\ 
When & \cjRL{mmty}/\cjRL{mty}  & 13.57\% \\ 
Where & \dots/\cjRL{hykn}/\cjRL{'yph} & 10.86\% \\ 
How & \cjRL{ky.sd}/\cjRL{'yk}  & 6.79\% \\ 
How much/many & \dots/\cjRL{bkmh}/\cjRL{kmh} & 5.43\% \\
Why & \cjRL{mdw`}/\cjRL{lmh} & 4.52\% \\
\bottomrule
\end{tabular}
\caption{Question word distribution, according to the first word of each question in the validation set. Inflected words and synonyms are clustered together to better align with English question types.}
\label{table:question-types}
\end{table}

\subsection{Question Diversity}

To measure the dataset's diversity, we cluster questions by their question word (typically the first word in the question).
Table~\ref{table:question-types} shows that \textit{what} (\cjRL{mh}) and \textit{which} (\cjRL{'yzh}) questions account for a third of the sample, with other answer types being distributed in a rather balanced distribution.
We also observe that about 11\% of the data contains \textit{how} (\cjRL{'yk}) and \textit{why} (\cjRL{lmh}) questions, which may reflect more complex instances.

\subsection{Sequence Length}

We measure the length in words (using whitespace tokenization) of each question and each answer.
Figure~\ref{fig:token-hist} shows the distributions of annotated questions and answers.
We observe that most questions use between 4-7 words, which is typical of simple questions in Hebrew.
More complicated questions constitute 27.6\% of the data, for example:
?\cjRL{'yk nqr't h'wprh h'.hrwnh /sktbw gylbr.t ws'lybn y.hdyw} (\textit{What is the last opera written jointly by Gilbert and Sullivan called?})
There are even questions with only 2 words; due to Hebrew's rich morphology, these questions are usually translated to 3-4 words in English, e.g. ?\cjRL{mhw hmnyk'yzm} (\textit{What is Manichaeism?})
Answer lengths, however, can vary greatly, depending on whether the annotators wrote minimal spans (typically 1-4 words) or included supporting information in the answer spans (see Section~\ref{sec:quality}).

\subsection{Linguistic Phenomena}

As a morphologically-rich language \cite{tsarfaty-etal-2010-statistical, seddah-etal-2013-overview}, modern Hebrew exhibits a variety of non-trivial phenomena that are uncommon in English and could be challenging for NLP models \cite{tsarfaty-etal-2020-spmrl}. We can identify some of these phenomena in our dataset.
Consider for example the following question-answer pair from the validation set:
\pex[numoffset=-2em, interpartskip=3ex, samplelabel=(a),labelformat=A:, avoidnumlabelclash]
\a[label=Q]
\begingl[everygla=\it,  belowglpreambleskip=0pt, aboveglftskip=0pt]
        \glpreamble
?\cjRL{mh hyh /s.t.hw /sl kpr /smryhw k/shwqm} //
        \gla ma haya shitkho shel kfar shmaryahu kshe-hukam //
        \glb what was area-of-it of Kfar Shmaryahu when-was.established //
        \glft `What was Kfar Shmaryahu's area when it was established?' //
\endgl

\a[label=A]
\begingl[everygla=\it, belowglpreambleskip=0pt, aboveglftskip=0pt]
        \glpreamble ... \cjRL{hy/swb hwqm `l /s.t.h /sl} //
        \gla ha-yeshuv hukam al shetakh shel ... //
        \glb the-village was.established on area of ... //
        \glft `The village was established on an area of ...' //
\endgl
\xe
This example illustrates a morphological variation between the question and the answer: the same entity appears as a morpheme in a compound word in the question's text: \cjRL{/s.t.hw} (\textit{its area}), \cjRL{k/shwqm} (\textit{when it was established}), but as a standalone word  (i.e. without inflection) in the answer: \cjRL{/s.t.h} (\textit{area}), \cjRL{hwqm} (\textit{was established}). These phenomena make exact match-optimized predictions more difficult for models aimed to solve this task.




\section{Baselines}

We establish baseline results for \textsc{ParaShoot} using BERT-style models.
Results indicate the task is challenging, leaving much room for future work in Hebrew NLP to advance the state of the art.

\subsection{Experiment Setup}

We fine-tune three adaptations of BERT \cite{devlin-etal-2019-bert}: 
\textit{mBERT}, trained by the original authors on a corpus consisting of the entire Wikipedia dumps of 100 languages;
\textit{HeBERT} \cite{chriqui2021hebert}, trained on the OSCAR corpus \cite{ortiz-suarez-etal-2020-monolingual} and Hebrew Wikipedia;
\textit{AlephBERT} \cite{seker2021alephbert}, also trained on the OSCAR corpus, with an additional 71.5 million tweets in Hebrew.
All models are equivalent in size to BERT-base, i.e. 12 layers, 768 model dimensions, and 110M parameters in total.

We fine-tune the models using the default implementation of HuggingFace Transformers \cite{wolf-etal-2020-transformers}. We select the best model by validation set performance over the following hyperparameter grid: learning rate $\in \{3\mathrm{e}{-5}, 5\mathrm{e}{-5}, 1\mathrm{e}{-4}\}$, batch size $\in \{16, 32, 64\}$, and update steps $\in \{512, 800, 1024\}$.
We compare the models' predictions to the annotated answer using token-wise F1 score and exact match (EM), as defined by \citet{rajpurkar-etal-2016-squad}.

\subsection{Results}
Table~\ref{table:results} shows the performance of each model on \textsc{ParaShoot}, with mBERT achieving the highest performance (56.1 F1).
We also observe significant variance across the models, with mBERT and AlephBERT performing significantly better than HeBERT.
It is not immediately clear where this discrepancy stems from; one possibility is that the introduction of noisy data via multilinguality (mBERT) or tweets (AlephBERT) makes that model more robust to potential noise in the annotated questions (e.g. typos).
Comparing these results to the estimated ceiling performance of 84 F1 (see Section~\ref{sec:quality}), we can infer that \textsc{ParaShoot} poses a genuine challenge to future Hebrew models and encourages further analysis of the semantic capabilities of the current models.

\begin{table}[t!]
\centering
\small
\begin{tabular}{@{}lcc@{}} 
\toprule   
\textbf{Model} & \textbf{F1} & \textbf{EM}  \\
\midrule
HeBERT & 36.7 & 18.2 \\ 
AlephBERT & 49.6 & 26.0\\
mBERT & \textbf{56.1} & \textbf{32.0} \\
\bottomrule
\end{tabular}
\caption{Baseline performance on the test set.}
\label{table:results}
\end{table}

\subsection{Error Analysis}
We analyze the error distribution by sampling 50 examples from the validation set and comparing AlephBERT's predictions to the annotated answers.
Table~\ref{table:error-analysis} shows how the examples are distributed into five categories, accounting for every type of overlap between the model's prediction and the annotated answer.
Putting aside exact matches (which account for about a quarter of examples), nearly half of the errors stem from zero overlap between the annotated answer and the model's prediction.
We observe that a significant part of the sample (22\%) contains cases where the annotated answer is a substring of the model's prediction, which might be, to a large extent, an artifact of the long answer annotations we observe in Section~\ref{sec:quality}.
For examples of erroneous predictions see Appendix~\ref{sec:appendix-b}.

\begin{table}[t!]
\centering
\small
\begin{tabular}{@{}lcc@{}} 
\toprule
\multirow{2}{*}{\textbf{Overlap Type}} & \textbf{Sample} & \textbf{Error} \\
 & \textbf{Frequency} & \textbf{Frequency} \\
\midrule
Model $=$ Annotation  & 26\% & -- \\
\midrule
Model $\subset$ Annotation  & 14\%                           & 19\%                      \\
Model $\supset$ Annotation & 22\%                           & 30\%                      \\
Model $\cap$ Annotation $\neq \emptyset$ & 4\%             & 5\%    \\
\midrule
Model $\cap$ Annotation $= \emptyset$ & 34\%                & 46\%    \\
\bottomrule
\end{tabular}
\caption{An error analysis of 50 random examples from the validation set, based on AlephBERT's predictions. The first reflects exact matches, and the last case accounts for zero overlap between model prediction and annotated answer.
The three categories in the middle refer to partially correct answers, where the model's prediction has some overlap with the annotated answer.}
\label{table:error-analysis}
\end{table}

\section{Conclusion}
In this paper, we present \textsc{ParaShoot}, the first question answering dataset in modern Hebrew, in a style and data collection methodology similar to that of SQuAD.
Baseline results demonstrate the potential of this dataset for researchers and practitioners alike to develop better models and datasets for natural language understanding in Hebrew.

\section*{Acknowledgements}
This work was supported by the Tel Aviv University Data Science Center, Len Blavatnik and the Blavatnik Family foundation, the Alon Scholarship, Intel Corporation, and the Yandex Initiative for Machine Learning. We thank Reut Tsarfaty for her valuable feedback.

\bibliographystyle{acl_natbib}
\bibliography{anthology,custom}

\appendix
\clearpage

\counterwithin{figure}{section}



\section{Error Examples}
\label{sec:appendix-b}

\begin{figure}[h]
    \centering
    \includegraphics[width=\textwidth,trim=2cm 6cm 1.5cm 2cm,clip]{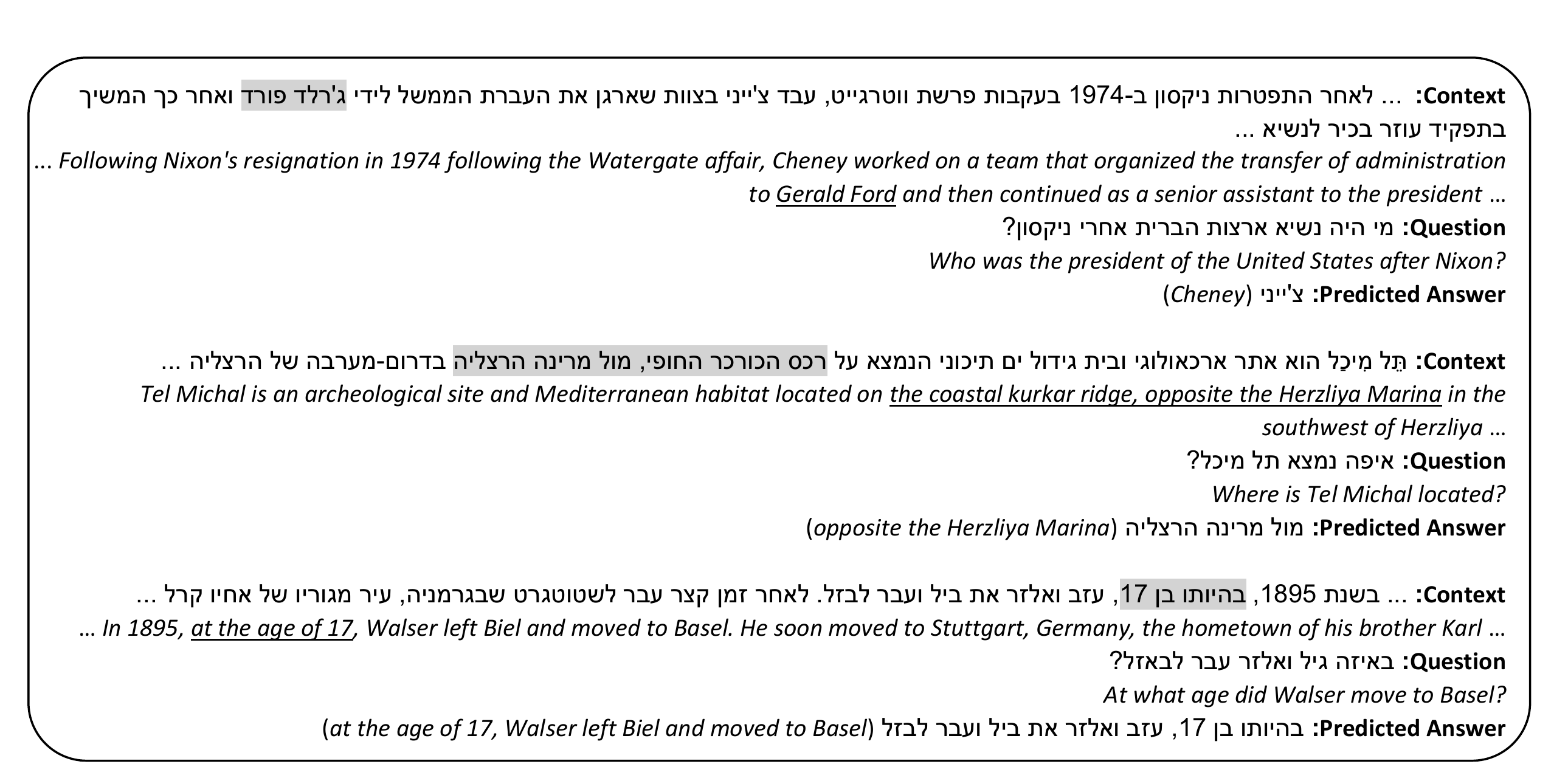}
    \caption{Predictions made by fine-tuned AlephBERT vs. annotated answers. In the first example, the prediction produced by the model is clearly an error. In the second example, the annotated answer span is excessively long, and the model predicts a more accurate substring of this span. In the third example, the model predicts a full sentence, while the annotated answer span is shorter.}
    \label{fig:error-example}
\end{figure}

\end{document}